\pdfoutput=1
\documentclass[11pt]{article}
\usepackage{acl}

\usepackage{times}
\usepackage{latexsym}
\usepackage[T1]{fontenc}
\usepackage[utf8]{inputenc}
\usepackage{microtype}
\usepackage{graphicx}
\usepackage{booktabs}
\usepackage{amsmath}
\usepackage{array}
\usepackage{enumitem}
\usepackage{tikz}
\usetikzlibrary{arrows.meta,positioning,fit,backgrounds}

\newcommand{\system}{\textsc{Muslim}}
\newcommand{\tool}[1]{\texttt{#1}}

\title{Muslim: A Deployed Arabic Voice AI Platform for Grounded Islamic Knowledge}

\author{Yahya Mohamed Elnawasany \\
  Independent Researcher, Egypt \\
  \texttt{yahyaalnwsany39@gmail.com}}

\begin{document}
\maketitle

\begin{abstract}
We present \system{}, a production Arabic voice AI platform serving grounded, sourced Islamic knowledge to real users. Beyond a real-time voice pipeline (NeMo Arabic ASR, an OpenAI-compatible LLM endpoint, self-hosted TTS) and a deterministic multi-source retrieval layer routed across six Model Context Protocol servers, we report three things a research prototype typically lacks. First, a released family of fine-tuned Arabic Islamic model artifacts: an efficient tool-routing LLM (Muslim-6B-PRO, 5.94B parameters) and a Modern Standard Arabic TTS model (Fasih-TTS-V1) that ranks 5th of 17 overall and 2nd of 11 open-weight systems on the community-voted Arabic TTS Arena for MSA. Second, an account and metering layer -- a free per-account turn allowance, capacity-aware refusal, and email verification deferred to the point it actually matters -- that turns an open demo into an operable, abuse-resistant product. Third, a three-layer observability stack (liveness, error reporting, product analytics) built specifically around the system's characteristic failure mode: a GPU-bound agent host going silent while the web tier keeps serving normally. We report real, measured latency and accuracy figures (98.4\% recitation-validation accuracy on 124 cases; end-to-end voice latency of 0.9--1.7s) and discuss the concrete engineering trade-offs and limitations of running an Islamic-knowledge voice product in production.
\end{abstract}

\section{Introduction}
\label{sec:intro}

Commercial voice assistants handle Arabic poorly, answer Islamic questions from parametric memory rather than cited texts, and route every conversation through third-party cloud infrastructure. For a domain where a misattributed narration is a real harm, these are not minor gaps. \system{} addresses them with an architecture built around three principles: Arabic (including Quranic Arabic) as a first-class citizen, no Islamic content generated without retrieval from a verified source, and the compute-heavy components of the pipeline running on infrastructure the operator controls rather than a third party.

This paper reports on \system{} as a \emph{deployed} system, not a prototype. Our contributions:

\begin{enumerate}[itemsep=1pt,topsep=2pt]
  \item A real-time Arabic voice pipeline integrating Arabic-specialized ASR, an env-configurable LLM endpoint, and self-hosted TTS, with a six-server tool-retrieval layer (\S\ref{sec:architecture}).
  \item A released family of fine-tuned Arabic Islamic model artifacts, including an MSA TTS model independently ranked on a community leaderboard (\S\ref{sec:models}).
  \item An account and metering layer -- a free rolling turn allowance, a fail-open capacity wall, deferred email verification -- that is, to our knowledge, undocumented in prior Islamic-voice-AI work because prior work did not need to survive real traffic (\S\ref{sec:accounts}).
  \item A three-layer observability design targeted at a specific, non-obvious deployed-system failure mode (\S\ref{sec:observability}).
  \item An evaluation combining real measured latency/accuracy figures with reliability evidence from an automated test suite (\S\ref{sec:eval}).
\end{enumerate}

\section{Related Work}
\label{sec:related}

Commercial voice assistants (Siri, Google Assistant, Alexa) answer Islamic questions from general-purpose parametric memory, without source attribution and without any Arabic-specific speech tuning. Text-based Islamic reference sites (Quran.com, Sunnah.com) provide searchable, sourced access to the Qur'an and Hadith but no conversational voice interface and no recitation feedback. Prior computational work on Quranic recitation has focused primarily on acoustic phoneme assessment via forced alignment or end-to-end classification \cite{radford2023whisper}, both of which require Tajweed-aware acoustic models or large labeled-error corpora that do not yet exist at production quality; \system{}'s validator instead operates on normalized ASR text, trading acoustic-level Tajweed detection for a deployable, deterministic pipeline with no training data requirement. On the generation side, retrieval-augmented generation \cite{lewis2020rag} is well established as a hallucination mitigation for open-domain QA; we apply it specifically to reference-addressable religious corpora, where the dominant query pattern names a specific verse or narration and a deterministic direct-lookup retriever is both simpler and more precise than approximate semantic search. To our knowledge, no prior published system combines real-time Arabic voice interaction, source-attributed multi-corpus retrieval, deterministic recitation validation, and a production account/metering/observability layer in one deployed platform.

\section{System Architecture}
\label{sec:architecture}

\begin{figure*}[t]
\centering
\begin{tikzpicture}[
  every node/.style={font=\small},
  box/.style={draw, rounded corners, minimum height=0.85cm, align=center, inner sep=3pt},
  toolbox/.style={draw, rounded corners, minimum height=0.7cm, align=center, font=\scriptsize, text width=2.1cm, inner sep=2pt},
  grouplabel/.style={font=\scriptsize\itshape},
  arr/.style={-{Latex[length=2mm]}, thick}
]
 \node[box, fill=blue!8, minimum width=1.8cm]           (browser) at (0,0)   {Browser};
 \node[box, fill=blue!8, minimum width=2.1cm]           (livekit) at (3,0)   {LiveKit SFU\\(WebRTC)};
 \node[box, fill=green!8, minimum width=3.4cm]          (agent)   at (7.2,0) {Voice Agent\\\scriptsize VAD $\to$ STT $\to$ LLM $\to$ TTS};

 \node[toolbox, fill=orange!10] (validator) at (5.4,-2.1)  {Validator\\\scriptsize(recitation)};
 \node[toolbox, fill=orange!10] (islamic)   at (7.7,-2.1)  {IslamicMCP\\\scriptsize(Tafsir, audio)};
 \node[toolbox, fill=orange!10] (hadith)    at (10.0,-2.1) {Hadith server\\\scriptsize(50k+ narrations)};

 \node[toolbox, fill=gray!10]   (tafsirnet) at (5.4,-3.9)  {Qur'anic\\metadata};
 \node[toolbox, fill=gray!10]   (websearch) at (7.7,-3.9)  {Web search};
 \node[toolbox, fill=gray!10]   (fiqhqa)    at (10.0,-3.9) {Fiqh Q\&A\\\scriptsize(optional)};

 \node[grouplabel, above=0.05cm of validator.north west, anchor=south west] {operator-controlled (local)};
 \node[grouplabel, below=0.05cm of tafsirnet.south west, anchor=north west] {external (remote HTTPS)};

 \draw[arr] (browser) -- (livekit);
 \draw[arr] (livekit) -- (agent);
 \draw[arr] (agent.south) -- ++(0,-0.5) -| (validator.north);
 \draw[arr] (agent.south) -- ++(0,-0.5) -| (islamic.north);
 \draw[arr] (agent.south) -- ++(0,-0.5) -| (hadith.north);
 \draw[arr] (agent.south) -- ++(0,-1.3) -| (tafsirnet.north);
 \draw[arr] (agent.south) -- ++(0,-1.3) -| (websearch.north);
 \draw[arr] (agent.south) -- ++(0,-1.3) -| (fiqhqa.north);
\end{tikzpicture}
\caption{System overview. The browser connects over WebRTC to LiveKit; the stateless agent runs the voice pipeline and dispatches tool calls to six MCP servers (Table~\ref{tab:mcp}) -- three operator-controlled, three external.}
\label{fig:architecture}
\end{figure*}
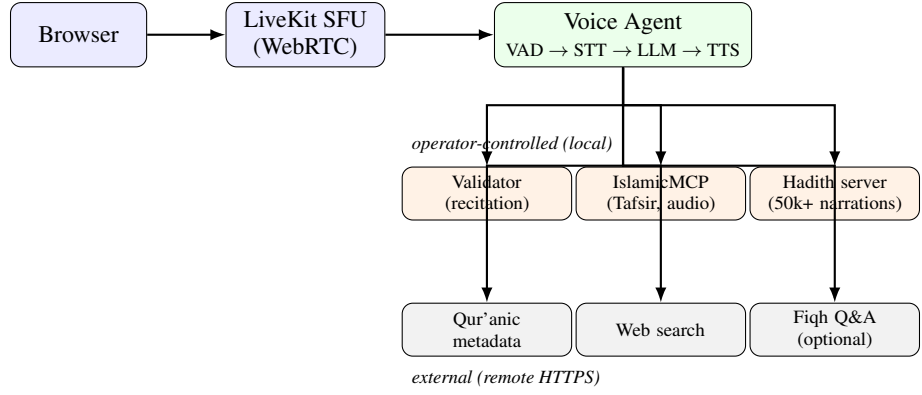

\system{} is a microservices architecture: a WebRTC selective forwarding unit (LiveKit; \citealp{livekit2024,livekit_agents2024}) carries audio between the browser and a stateless voice agent, which runs Silero VAD \cite{silero_vad2021}, an Arabic ASR model, an LLM, and TTS, and dispatches tool calls to knowledge servers over the Model Context Protocol (MCP; \citealp{mcp2024}) via FastMCP \cite{fastmcp2024}. The frontend (Next.js; \citealp{nextjs2024}) mints the only credential that lets a client reach the agent, and is where account/metering logic lives (\S\ref{sec:accounts}). Figure~\ref{fig:architecture} shows the complete system.

\subsection{Voice Pipeline}
\label{sec:voicepipeline}
ASR uses NVIDIA NeMo's \cite{kuchaiev2019nemo} Arabic FastConformer model \cite{rekesh2023fastconformer}, chosen over general-purpose alternatives because its training distribution includes formal and religious Arabic, which matters directly for Quranic vocabulary that general web-audio-trained models under-represent. VAD is preloaded once per worker process to eliminate cold-start latency, and the session uses preemptive generation -- beginning LLM inference during the trailing-silence window VAD requires before confirming end-of-utterance -- to overlap two otherwise-sequential latency sources. The LLM is reached through a single OpenAI-compatible interface, keeping the serving backend a two-variable configuration choice (endpoint URL, model name) rather than a code dependency; two small startup patches correct tool-schema and tool-call-token serialization quirks specific to the current OpenAI-compatible backend, without modifying the agent SDK itself. TTS is selected the same way, defaulting to the self-hosted model in \S\ref{sec:models} with a cloud fallback available behind the same switch.

\subsection{Deployment: Two Coexisting Topologies}
\label{sec:deployment}
The system supports two deployment modes on identical application code, differing only in \emph{where} each service runs. A single-host, fully self-hosted mode runs every service in containers on one operator-controlled machine. Our own production deployment instead runs a \emph{hybrid} topology: the web/account tier runs on a managed platform with a managed Postgres database, the WebRTC media layer runs on a managed real-time infrastructure provider, and the voice agent together with the GPU-dependent ASR/TTS services run on a single operator-controlled host that \emph{dials out} to the media layer and accepts no inbound connections. This keeps the Arabic-specific, compute-heavy core of the system entirely under operator control -- the part this paper's contributions concern -- while the web tier benefits from managed-platform availability. The trade-off, discussed in \S\ref{sec:limitations}, is that the web tier and the voice agent can now fail independently: the website can be fully up while no conversation is possible.

\subsection{Knowledge Retrieval: Six MCP Servers}
\label{sec:retrieval}
Islamic knowledge is retrieved, never generated from parametric memory. Table~\ref{tab:mcp} lists the six MCP servers the agent connects to; three run on operator-controlled infrastructure. Tafsir (Quranic exegesis) is served by direct JSON file lookup across eight classical Arabic books -- no embedding model, no cold start, sub-millisecond latency, exact precision on direct-reference queries. Hadith retrieval, previously delegated to an external community service that proved unreliable in practice, is now served by a self-hosted server over 50,000+ narrations across 17 collections, removing a third-party dependency from the critical path. Quranic audio is never synthesized by TTS: a verse request bypasses the LLM/TTS branch entirely and streams an authentic human recitation, because no TTS system reliably reproduces Tajweed phonological rules and recitation is, in the tradition this system serves, a specialized discipline with abundant authentic recordings already available.

\begin{table}[t]
\centering
\small
\begin{tabular}{p{2.1cm}p{1.1cm}p{3.3cm}}
\toprule
\textbf{Server} & \textbf{Locus} & \textbf{Role} \\
\midrule
Validator & local & Recitation validation (\S\ref{sec:eval}) \\
IslamicMCP & local & Tafsir lookup, audio URLs \\
Hadith server & local & 50k+ narrations, 17 collections \\
Qur'anic metadata & remote & Revelation context, qira'at, morphology \\
Web search & remote & Contemporary topics, neural + keyword \\
Fiqh Q\&A & remote & Scholarly Q\&A corpus (optional) \\
\bottomrule
\end{tabular}
\caption{The six MCP servers the agent can call. Three run on operator-controlled infrastructure; three are external services reached over HTTPS with no operator data retained.}
\label{tab:mcp}
\end{table}

\section{Released Model Family}
\label{sec:models}

Rather than depending solely on third-party inference, we release a family of fine-tuned Arabic Islamic model artifacts, available at \url{https://huggingface.co/NightPrince}. Table~\ref{tab:models} summarizes the two primary releases.

\begin{table}[t]
\centering
\small
\begin{tabular}{p{1.6cm}p{5.3cm}}
\toprule
\textbf{Model} & \textbf{Summary} \\
\midrule
Muslim- \newline 6B-PRO & QLoRA fine-tune (4-bit NF4, $r{=}16$) \cite{dettmers2023qlora,hu2022lora} of a 5.94B-parameter Qwen3-architecture base \cite{qwen3_2025}, 262K context, on 2{,}731 curated tool-calling / production-derived examples via TRL \cite{trl2020}. Built for tool-call routing (31 tools) and persona discipline, not unsupported scripture recitation. Apache-2.0 \cite{muslim6bpro}. \\
\addlinespace
Fasih- \newline TTS-V1 & Fine-tune of Coqui XTTS~v2 \cite{casanova2024xtts} on 1{,}297 curated single-speaker MSA clips ($\sim$2.4h), with a custom Arabic front-end (normalization, diacritization, chunking). CER 1.3\% (vs. 1.8\% human-recording baseline) on the SILMA Open-Source Arabic TTS Benchmark \cite{silmattsbenchmark}, ranking highest for intelligibility among open-source systems tested. Coqui Public Model License (non-commercial) \cite{fasihttsv1}. \\
\bottomrule
\end{tabular}
\caption{The two primary released model artifacts. Both are used in the production pipeline behind an environment-switchable configuration; the LLM currently serving live conversation traffic is a larger general-purpose model reached through the same OpenAI-compatible interface, with Muslim-6B-PRO as the path toward fully self-hosted inference.}
\label{tab:models}
\end{table}

We independently verified Fasih-TTS-V1's standing on the Arabic TTS Arena \cite{arabicttsarena2026}, a community-voted, ELO-rated leaderboard, by retrieving its underlying results data directly rather than relying on a rendered page. As of our snapshot (2026-08-18; 245 head-to-head MSA battles), Fasih-TTS-V1 ranks \textbf{5th of 17} systems overall on Modern Standard Arabic and \textbf{2nd of 11} among open-weight systems, ahead of several commercial entrants. Because this is a live, continuously updated arena, we report the snapshot date and battle count rather than treating the rank as fixed. Two further fine-tuned Arabic ASR checkpoints (a Quran-recitation-specialized FastConformer model and a diacritization-aware NeMo model) round out the released family but are not evaluated further in this paper.

\section{Accounts, Metering, and the Capacity Wall}
\label{sec:accounts}

A single server route mints the only credential that lets a client join a session with the agent, and is therefore where every access-control decision is concentrated. It resolves the caller's session, checks their entitlement, and either refuses with one of six typed reasons (\tool{no\_session}, \tool{signin\_required}, \tool{quota\_exhausted}, \tool{verify\_email\_required}, \tool{at\_capacity}, \tool{agent\_offline}) or signs the allowance directly into the issued token as participant attributes the client cannot forge.

\textbf{Tiers.} An account is the only way into a conversation: there is no guest tier and no turns before sign-up. The service is free to every user and there is no paid tier: the single remaining allowance grants 30 turns on a \emph{rolling 24-hour} window, a ceiling set by the fixed GPU capacity the deployment runs on rather than by a pricing model. This is a reversal. The system shipped with an anonymous tier carrying a small lifetime allowance, which acted as a preview and a sign-in wall, and it was retired in September 2026 after the free-allowance question was settled the other way. The migration is the part worth reporting: guest rows were kept rather than deleted, because their conversations carry captured turns that a cascading delete would have taken with them, and the session layer instead reads a surviving guest cookie as signed out. Enforcement is layered: the agent itself counts turns against the limit signed into its token and closes the session on reaching it -- a hard stop that holds even if every usage report to the database fails -- while the database accumulates cross-session usage for the rolling window, counted by turn index so a retried report never double-counts a user.

\textbf{Deferred verification.} Email confirmation is required to \emph{refill} the allowance, not to sign up. Sign-up and first use are the highest-intent moment a visitor has, and putting a clicked email link between that intent and the product spends it; withholding only the refill defers the cost to a point where the user already knows whether the product is worth the click. An unverified account is therefore worth exactly one window's turns. Retiring the guest tier widened this rather than narrowing it: with no lifetime allowance left, the grace period is now the first wall every user meets rather than the second.

\textbf{Capacity wall.} Because the GPU-dependent services run on a fixed-size host, the token route also checks the agent's most recently reported active-job count before issuing a token, refusing with a distinct reason when saturated rather than letting the user experience an unexplained failed join -- the two are otherwise indistinguishable from the user's side. This check is deliberately \emph{fail-open}: if the capacity query itself errors, requests are allowed through, so a monitoring fault cannot itself take the product down.

\section{Observability}
\label{sec:observability}

\system{}'s characteristic failure mode defeats naive monitoring: the GPU-bound agent host can go silent while the web application keeps serving 200s normally, so pinging the website learns nothing. We address this with a heartbeat that travels \emph{outward} from the agent on a fixed interval to a judge that sits outside both halves of the system; staleness alone reveals an outage, with no crash-detection logic required. A public health endpoint encodes a three-way verdict (healthy / degraded -- notably, alive but not registered with the media layer, invisible to a naive check / down) in both its body and HTTP status.

This liveness signal is deliberately one of three independent layers. Error reporting captures both web-tier exceptions and agent-side failures the agent deliberately swallows to keep a conversation alive -- exactly the failures a healthy heartbeat cannot reveal. Product analytics tracks a deliberately small set of funnel events (session start, refusal reason, sign-up, verification, auth failure, and support contact), counting a conversation as started only when a token is actually issued rather than when a button is pressed, since the button is also pressed by users about to be refused. Privacy choices are consistent across all three layers: no session replay, no default PII capture, and analytics identity keyed to an internal id, never an email address.

\section{Evaluation}
\label{sec:eval}

\subsection{Recitation Validator}
The deterministic recitation validator -- a seven-step Arabic normalization pipeline followed by a four-layer verse-matching search, with no LLM involvement -- scores \textbf{98.4\%} (122/124) on the 124-case suite released with it. The validator, its 2{,}290-pair Uthmani-to-Standard word mapping, and that evaluation are the subject of a companion paper \cite{elnawasany2026uthmani}, which we cite rather than restate; the figures here are quoted from the released artifact and reproduce by running it.

Two results from that work bear on this system. First, both failures share one mechanism -- a single substitution error can make a \emph{different} verse a literal exact match, which is a property of the search design rather than of the morphological data. Second, a corpus-wide census finds that 16.5\% of the Qur'an's 6{,}236 verses (369 groups) share an ambiguous four-word normalized opening with at least one other verse; the remaining 83.5\% resolve unambiguously. The ambiguity is a property of the text, not a pipeline defect, and it is the reason the product asks for continuation rather than guessing between candidates.

One measurement result is worth reporting at this venue in its own right, because it is an operational failure rather than a modeling one. An earlier figure of 99.2\% for this same validator was obtained in an environment where an optional fuzzy-matching dependency happened to be absent. The code logged a warning and returned an empty candidate list, so the fourth search layer was silently skipped rather than failing, and the evaluation scored a three-layer system while the documentation described four. The released artifact now depends on nothing outside the Python standard library, which removes the class of error rather than the instance: an accuracy number that moves with which packages are installed is not a property of the system being measured.

\subsection{Latency}
Measured directly on real data: ASR mean latency 235.5ms (RTF 0.025, i.e. real audio is transcribed roughly 40$\times$ faster than it plays), LLM first-token mean latency 519.7ms. Estimated end-to-end latency from speech end to first audio response is 0.9--1.7s under typical conditions -- competitive with commercial assistants despite running the ASR/TTS stages on operator-controlled rather than hyperscale infrastructure.

\subsection{Retrieval Grounding}
On 30 held-out Islamic questions, retrieval succeeds for 100\% of queries, and the retrieval-augmented condition adds only $\sim$50ms mean latency over answering from parametric memory alone (786ms vs. 736ms) -- confirming that grounding, which is the paper's central integrity claim, is not purchased at a meaningful latency cost.

\subsection{Reliability Evidence}
Beyond point-in-time accuracy, an automated suite of over 1{,}000 tests (889 Python, 119 TypeScript) covering the validator, retrieval tools, and account/metering logic runs on a pre-push git hook mirroring a configured continuous-integration workflow; the CI workflow itself is currently inactive pending platform billing resolution on our private repository, a mundane operational constraint we report because it is the kind of practical deployment friction this venue specifically solicits, not because we consider it resolved.

\section*{Limitations}
\label{sec:limitations}

\textbf{Single-host availability.} The hybrid topology (\S\ref{sec:deployment}) means the web tier and account system remain available independently of the voice agent, but no conversation is possible while the single GPU host is offline -- acceptable during beta, not for a paying userbase without redundant capacity. \textbf{Dialectal Arabic.} The ASR model targets Modern Standard and Quranic Arabic; dialectal speakers may see elevated error rates. \textbf{No formal user study.} All results are technical (accuracy, latency, retrieval success); no controlled study of learning outcomes or perceived response quality has been conducted. \textbf{RAG coverage boundary.} Questions outside the indexed Tafsir/Hadith/metadata corpora fall back to the LLM's parametric knowledge and may be inaccurate; a formal faithfulness evaluation (e.g. RAGAS; \citealp{es2023ragas}) against expert-annotated ground truth is future work. \textbf{Model-family evaluation.} Muslim-6B-PRO's model card does not yet report quantified benchmark results; we describe its architecture and training honestly rather than assert an accuracy figure it does not have. \textbf{Live-leaderboard citation.} The Arabic TTS Arena rank reported in \S\ref{sec:models} reflects a snapshot of a continuously updated, community-voted system and may shift.

\section*{Ethical Considerations}
\label{sec:ethics}

The system captures voice conversations for service improvement; consent is signed into the same access token that carries identity, so recording honors a user's choice independent of whether usage metering is enabled, and an operator-level switch and the user's own choice are both required for capture to occur. No session replay is enabled anywhere in the stack, and product-analytics identity is keyed to an internal identifier, never an email address -- a deliberate choice given that a compromised analytics record for this product is a record of someone's private religious inquiry. Because Islamic knowledge questions carry real consequences when answered wrongly, every retrieval path is designed to make the system decline or hedge rather than answer confidently from unverified parametric memory outside the domains explicitly covered by the retrieval layer (\S\ref{sec:limitations}).

\section*{Acknowledgments}

The author thanks Prof.\ Marwa Seddiq for supervision and guidance throughout this project.

\bibliography{references_systems}

\end{document}